\documentclass{article}
\usepackage[T1]{fontenc}
\usepackage[utf8]{inputenc}
\usepackage{array}
\usepackage{amsmath}
\usepackage{amsthm}
\usepackage{amssymb}
\usepackage{graphicx}
\PassOptionsToPackage{normalem}{ulem}
\usepackage{ulem}

\makeatletter

\providecommand{\tabularnewline}{\\}

\usepackage[numbers]{natbib}
\usepackage[final]{nips_2017}

\usepackage{url}
\usepackage{amsfonts}
\usepackage{nicefrac}
\usepackage{microtype}

\usepackage{graphicx}

\title{Predicting Movie Genres Based on Plot Summaries}

\author{
   Quan Hoang\\
   University of Massachusetts-Amherst\\
  \texttt{qhoang@umass.edu} \\
}
\usepackage{graphicx}

\usepackage{colortbl} 
\definecolor{header_color}{rgb}{0.74,0.88,0.91}
\definecolor{even_color}{rgb}{0.9,0.9,0.9}
\definecolor{subheader_color}{rgb}{0.85,0.93,0.95}
\definecolor{childheader_color}{rgb}{1.0,0.93,0.87}

\definecolor{ccolor_best}{rgb}{1.0,0.9,0.9}
\definecolor{ccolor_wrong}{rgb}{1.0,0.85,0.85}

\usepackage{ifthen}
\renewenvironment{figure}[1][]{%
 \ifthenelse{\equal{#1}{}}{%
   \@float{figure}
 }{%
   \@float{figure}[#1]%
 }%
 \centering
}{%
 \end@float
}

\renewenvironment{table}[1][]{%
 \ifthenelse{\equal{#1}{}}{%
   \@float{table}
 }{%
   \@float{table}[#1]%
 }%
 \centering
}{%
 \end@float
}

\usepackage{longtable}

\makeatother

\begin{document}

\title{Predicting Movie Genres Based on Plot Summaries}
\maketitle
\begin{abstract}
This project explores several Machine Learning methods to predict
movie genres based on plot summaries. Naive Bayes, Word2Vec+XGBoost
and Recurrent Neural Networks are used for text classification, while
K-binary transformation, rank method and probabilistic classification
with learned probability threshold are employed for the multi-label
problem involved in the genre tagging task. Experiments with more
than 250,000 movies show that employing the Gated Recurrent Units
(GRU) neural networks for the probabilistic classification with learned
probability threshold approach achieves the best result on the test
set. The model attains a Jaccard Index of 50.0\%, a F-score of 0.56,
and a hit rate of 80.5\%.
\end{abstract}

\section{Introduction\label{sec:Introduction}}

Supervised text classification is a mature tool that has achieved
great success in a wide range of applications such as sentiment analysis
and topic classification. When applying to movies, most of previous
work has been focused on predicting movie reviews or revenue, and
few research was done to predict movie genres. Movie genres are still
tagged through a manual process in which users send their suggestions
to email address of The Internet Movie Database (IMDB). As a plot
summary conveys much information about a movie, I explore in this
project different machine learning methods to classify movie genres
using synopsis. I first perform experiment with Naive Bayes using
bag-of-word features. Next, I make use of the pretrained word2vec
embeddings \cite{mikolov2013efficient,pretrained2013wordtovec} to
turn plot summaries into vectors, which are then used as inputs for
an XGBoost classifier \cite{chen2016xgboost}. Finally, I train a
Gated Recurrent Unit (GRU) neural network \cite{cho2014properties}
for the genre tagging task.

The rest of the report is organized as follows. Section~\ref{sec:Related-work}
discusses the related work. Section~\ref{sec:Data} describes the
dataset used for this project. Section~\ref{sec:Methodology} outlines
the models for experiments. Section~\ref{sec:Experiment} presents
experiment results. Finally, Section~\ref{sec:Conclusion} summarizes
the paper and discusses directions for future work.

\section{Related work\label{sec:Related-work}}

\citep{makita2016multinomial} proposes a Naive Bayes model to predict
movie genres based on user ratings of the movie. The idea is that
users are usually consistent with their preference and prefer some
genres over the others. The evaluation metric is the percentage of
movies that the model predicts correctly at least one of the true
labels. The focus of my project is natural language processing, so
I attempt to predict movie genres using only movie plot summary. Moreover,
I adopt more rigorous metrics such as F-score or Jaccard index. 

\citep{blackstock2008classifying} attempts to classify movie scripts
by building a logistic regression model using NLP-related features
extracted from the scripts such as the ratio of descriptive words
to nominals or the ratio of dialogues frames to non-dialogue frames.
For each movie scripts, the model, based on extracted features, estimates
the probability that the movie belong to each genre and takes the
$k$ best scores to be its predicted genres, where $k$ is a hyper-parameter.
The experiment is done on a small dataset with only 399 scripts and
the best subset of features achieves an F1 score of 0.56. 

\citep{Ho2011movies} investigates different methods to classify movies'
genres based on synopsis. The methods examined include One-Vs-All
Support Vector Machines (SVM), Multi-label K-nearest neighbor (KNN),
Parametric mixture model (PMM) and Neural network. All these methods
use the term frequency inverse document frequency of the words as
features. The dataset used for experiment is relatively small with
only 16,000 movie titles for both the train and test sets. In addition,
the experiment is limited to only predicting only 10 most popular
genres, including action, adventure, comedy, crime, documentary, drama,
family, romance, short films, and thrillers. Overall, SVM achieves
the highest F1 score of 0.55.

As a movie can belong to several genres, this project is related to
the multi-label classification problem, which was examined in some
previous work. \citep{schapire2000boostexter} introduces two extensions
of the AdaBoost algorithm \citep{freund1995desicion} for multi-class,
multi-label text categorization. The first extension turns the multi-label
problem into multiple, independent binary classification problems,
while the second ranks the labels so that the correct labels receives
the highest ranks. Both extensions are trained using weak hypotheses
as one-level decisions trees that check the presence or absence of
a term in a given document. For the second extension, the objective
was to minimize the average fraction of pairs of labels that are misordered,
a quantity called the empirical rank loss:

\begin{equation}
\frac{1}{m}\frac{1}{\vert Y_{l}\vert\vert\mathcal{Y}\setminus Y_{l}\vert}\vert\left\{ \left(i,j\right)\in Y_{l}\times\left(\mathcal{Y}-Y_{l}\right):r_{i}\left(\mathbf{x}_{l}\right)\leq r_{j}\left(\mathbf{x}_{l}\right)\right\} \vert
\end{equation}
where $\mathcal{Y}$ is the set of all label, $\mathbf{x}_{l}$ and
$Y_{l}$ are respectively the feature vector and the set of true labels
for the $l$-th example, $r_{i}\left(\mathbf{x}\right)$ is the rank
value function for label $i$, and $m$ is the total number of training
examples. The predicted labels are those with the top $k$ highest
ranks, where $k$ is a hyperparameter. Trained and tested on some
multi-label subset of the Reuter-21450 dataset, the rank-based extension
achieves slightly worse performance.

Inspired by the ranking method in \citep{schapire2000boostexter},
\citep{elisseeff2002kernel} propose a kernel method to learn the
ranking function. Instead of fixing a number $k$ as the label size
for prediction, this method also learn a threshold $t\left(\mathbf{x}_{l}\right)$.
More specifically, for an input vector $\mathbf{x}_{l}$ and the rank
values $\left(r_{1}\left(\mathbf{x}_{l}\right),r_{2}\left(\mathbf{x}_{l}\right),...,r_{Q}\left(\mathbf{x}_{l}\right)\right)$
where the set of labels is $\mathcal{Y=}\left\{ 1,2,...,Q\right\} $,
the threshold value is defined as:

\begin{equation}
t\left(\mathbf{x}\right)=\arg\min_{t}\vert\left\{ k\in Y_{l}\right\} \ s.t.\ r_{k}\left(\mathbf{x}_{l}\right)\leq t\vert+\vert\left\{ k\in\left(\mathcal{Y}\setminus Y_{l}\right)\right\} \ s.t.\ r_{k}\left(\mathbf{x}_{l}\right)\geq t\vert\label{eq:rank_threshold}
\end{equation}

When the minimum is not an unique value but a segment, the threshold
is chosen to be the middle of the segment. The threshold is modeled
using linear least squares.

\citep{zhang2006multilabel} applies this idea to neural networks
by slightly modify the rank loss function as:

\begin{equation}
\frac{1}{m}\frac{1}{\vert Y_{l}\vert\vert\mathcal{Y}\setminus Y_{l}\vert}\sum_{i\in Y_{l},\,j\in\left(\mathcal{Y}\setminus Y_{l}\right)}\exp\left(-\left(r_{i}\left(\mathbf{x}_{l}\right)-r_{j}\left(\mathbf{x}_{l}\right)\right)\right)\label{eq:rank_loss_function}
\end{equation}

The use of the exponential function is to severely punish the model
for assigning higher rank value to the wrong labels.

This project employs several machine learning methods to predict movie
genres, including Naive Bayes, XGBoost, and recurrent neural network
(RNN). XGBoost \citep{chen2016xgboost} is an advanced and efficient
implementation of the gradient boosting algorithm \citep{breiman1997arcing,friedman2001greedy,friedman2002stochastic},
an ensemble method that sequentially add new predictors that are trained
on the residual errors made by previous predictors. Unlike the standard
gradient boosting algorithm, XGBoost has a regularized objective and
and it makes use of second-order approximation to quickly and greedily
fit a new predictor at each iteration. XGBoost was the algorithm of
choice winning solutions of many machine learning and data mining
challenges.

To extract features to be used in XGBoost, I make use of the word2vec
framework proposed in \citep{mikolov2013efficient}, which learns
high-dimensional word embeddings. Word2vec learns embedding by training
a neural network to predict neighboring words. The idea is that semantically
similar words tend to occur near each other in text, so embeddings
that are good at predicting context words are also good at representing
similarity. There are two model architectures in word2vec to learn
word embeddings: continuous bag-of-words (CBOW) and continuous skip-gram.
In the CBOW architecture, the model predicts the current word from
a window of surrounding context words. In the continuous skip-gram
architecture, the model predicts the surrounding window of context
words using the current word. Vector embeddings learned by word2vec
was shown to capture relations between words. For example, the result
of the embedding vector('\emph{king}') - vector('\emph{man}') + vector('\emph{woman}')
is a vector close to vector('\emph{queen}'). For this project, I use
the pretrained 300-dimensional embeddings trained on part of the Google
News corpus for 3 billion words \citep{pretrained2013wordtovec}.
The average of the embedding vectors of all words in a movie plot
is then used as the vector representation of the plot.

Taking average of the embedding vectors of all words in a plot might
be a lossy summary and fail to capture the sequential relationship.
Therefore, I consider Recurrent Neural Networks (RNN), which are a
powerful family of neural networks for processing sequential data.
RNNs take one input from the input sequence at each time step. Hidden
units in RRNs take inputs not only from the data or from previous
layers as in traditional neural networks, but also from themselves
in the previous time step. This allows RNN units to take information
from the past sequence of inputs. Parameters in RNNs are shared across
time steps and learned by gradient descent optimization methods with
the Back Propagation Through Time (BPTT) algorithm \citep{werbos1988generalization},
which computes the gradients of the loss function with respect to
the parameters.

Gradients in RNNs, however, tend to vanish or explode as explored
in depth in \citep{hochreiter1991untersuchungen,bengio1994learning},
making it difficult to train RNNs. The LSTM networks \citep{hochreiter1997long}
address this problem by introducing a complex unit called \emph{memory
cell}. The central feature of LSTM's memory cell is the constant error
carrousel (CEC), which a self-connection for each cell state, thus
allowing the gradient signal to stay constant as it flows backward
across time steps. As memory cells interract with each other, they
are equipped with \emph{input gates} and \emph{output gates} to protect
themselves from perturbation. These input and output gates use multiplicative
operations to control the network's sensitivity to each unit's inputs
(or outputs).

Cell states in the original LSTM model, however, could become very
large and saturate the output gates, thus leading to gradient vanishing.
This problem was termed \textquotedblleft \emph{internal state drift}\textquotedblright{}
\citep{hochreiter1997long}. To tackle this issue, \citep{gers1999learning}
introduced \emph{forget gates}, which reset cell states when the gates
decide that the cell states' content is out-of-date. \textquotedblleft Reset\textquotedblright{}
does not mean to set the cell states to zero immediately, but to gradually
reset by multiplying with a number between 0 and 1. This architecture
allows the gradients to flow for long duration, and the LSTM has been
found extremely successful in many applications such as speech recognition
or machine translation.

Many LSTM variants have been proposed. Among them, Gated Recurrent
Unit (GRU) \citep{cho2014properties} has become increasingly popular.
GRU is similar to LSTM but combines the forget and input gates into
a single update gate. As GRU has a simpler architecture than LSTM,
I will use this architecture for the experiments with RNNs. In addition,
I experiment with a variant of GRU that I term Self-normalizing GRU
(SNGRU). In a GRU cell, the short-term memory states are squashed
to the range between -1.0 and 1.0 by the hyperbolic tangent (\emph{tanh})
function before going through the update gates. As a result, gradient
signals become weaker when flowing through update gates during backward
passes. The motivation of SNGRU is to avoid using the squashing function.
To keep cell states bounded, SNGRU applies layer normalization \citep{ba2016layer},
which helps the cell states converge to a normal distribution.

\section{Data\label{sec:Data}}

To prepare the dataset for experiments, I extract the movie plot summaries
and their corresponding set of genres from the IMDB datasets. The
raw data files, including the plot list and the genre list files,
are downloaded from \cite{imdb2017data}. The genre list file has
1,433,423 pairs of movie name and genre. It is noteworthy that there
are $k$ pairs with the same movie name in the file if the movie belongs
to $k$ genres. The plot list file has 345,920 pairs of movie and
plot summary. There are 338,789 movies for which both genres and plot
summaries are available. Selecting only movies belong to at least
one of the 20 genres, which will be listed below, results in 283,355
movies. Finally, filtering out movies with more than 200 tokens in
their plot summaries leaves 255,853 movies. I divide the data into
train and test sets according to the 80/20 proportion. As a result,
there are 204,682 movies in the train data, and 51,171 movies in the
test data. There are 15,187,708 tokens and 266,084 unique word types
in all of the train plot summaries. Of these, 60,325 word types occur
5 times or more.

The genre names and the percentages of movies in them are: \emph{drama}
(46.0\%), \emph{comedy} (27.9\%), \emph{thriller} (11.2\%), \emph{romance}
(11.0\%), \emph{action} (9.7\%), \emph{family} (8.1\%), \emph{horror}
(7.7\%), \emph{crime} (7.3\%), \emph{adventure} (7.0\%), \emph{animation}
(6.2\%), \emph{fantasy} (5.9\%), \emph{sci-fi} (5.6\%), \emph{mystery}
(5.4\%), \emph{biography} (5.2\%), \emph{music} (4.9\%), \emph{history}
(4.7\%), \emph{war} (2.8\%), \emph{western} (2.4\%), \emph{sport}
(2.4\%) and \emph{musical} (2.2\%).

In the train dataset, there are 4,145 unique sets of genres. Each
movie belongs on average to 1.84 genres. The distribution of the number
of genres a movie belong to are: 1 (50.5\%), 2 (25.2\%), 3 (17.1\%),
4 (5.1\%), 5 (1.5\%) and 6 or higher (0.6\%).

Of all the movie plot summaries, 11\% have 50 or fewer tokens, 56\%
have between 50 and 100 tokens, 25\% have between 100 and 150 tokens,
and 8\% has between 150 and 200 tokens. The average tokens per plot
summary is 89.5.

\section{Methods\label{sec:Methodology}}

\subsection{Naive Bayes}

For this task, I experiment with three machine learning methods. The
first method is to use Naive Bayes, a simple approach that was shown
very effective for many text classification tasks. Naive Bayes makes
two key assumptions. The first is the bag-of-words assumption, meaning
that the order of words does not matter. The second is the conditional
independence assumption. For document $\mathbf{d}=w_{1}w_{2}...w_{N}$
and given the class $c$, the probabilities $P\left(w_{i}\vert c\right)$
are independent. Thus:

\begin{equation}
P\left(w_{1},w_{2},...,w_{n}\vert c\right)=\prod_{i=1}^{N}P\left(w_{i}\vert c\right)
\end{equation}

The equation for the class chosen by a Naive Bayes classifier is thus:
\begin{align}
c_{NB} & =\arg\max_{c\in C}\frac{p\left(c\right)\prod_{i=1}^{N}P\left(w_{i}\vert c\right)}{P\left(\mathbf{d}\right)}\label{eq:NB_conditional_prob}\\
 & =\arg\max_{c\in C}p\left(c\right)\prod_{i=1}^{N}P\left(w_{i}\vert c\right)
\end{align}

I consider two approaches to apply Naive Bayes to the multi-label
classification task involved in this project. The first approach is
to build a binary Naive Bayes classifier (bNB) for each genre. The
second approach is to build a multinomial Naive Bayes classifier (mNB)
model. For each plot $\mathbf{d}$, MNB can estimate the posterior
probability $p\left(c\vert\mathbf{d}\right)$ in the right hand side
of Eq.~\ref{eq:NB_conditional_prob}. The model predicts that a movie
belong to a genre $c$ if $p\left(c\vert\mathbf{d}\right)>p\left(c\right)$,
where $p\left(c\right)$ is the prior probability of the genre $c$.
The intuition is that the model assigns a movie to a genre if the
model has a stronger than prior belief in the presence of the genre
after reading the movie's plot summary.

\subsection{Word2Vec$+$XGBoost\label{sub:Word2Vec-+-XGBoost}}

One problem with Naive Bayes is that the bag-of-word feature as well
as the conditional independence assumption are not realistic and may
not capture the meaning of sentences in text. Therefore, the second
method is to employ rich, contextual representations for words learned
by the word2vec framework \cite{mikolov2013efficient}. Each plot
summary is turned into a vector representation by taking average of
the embedding vectors of all words in each the text. The assumption
is that the average vector can be a good semantic summarization of
the plot summary.

The plot vector representation can then be fed into XGBoost \cite{chen2016xgboost}.
I apply XGBoost to the multi-label classification task in a similar
spirit to the ranking method \cite{elisseeff2002kernel} discussed
in Section~\ref{sec:Related-work}. Instead of learning the rank
values, however, I train an XGBoost classifier that output the probabilities
$P\left(c\vert\mathbf{x}\right)$ that the movie represented by $\mathbf{x}$
belongs to the genre $c$. To make predictions based on these probabilities,
a XGBoost regressor is trained to predict the probability threshold
for each categorical distribution.

To train the XGBoost classifier, the train data is modified by turning
each pair of plot vector and its corresponding set of correct genres
$\left(\mathbf{x},C\right)$ into $k$ pairs of $\left(\mathbf{x},c\right)$,
where $k=\vert C\vert$ and $c\in C$. To train the XGBoost regressor,
the probabilities $P\left(c\vert\mathbf{x}\right)$ estimated by the
XGBoost classifier for each movie in the train data are used as features,
and the probability threshold computed based on the corresponding
set of correct genres is used as the target. More specifically, the
probability threshold is defined as:

\begin{equation}
t\left(\mathbf{x}\right)=\arg\min_{t}\vert\left\{ c\in C\right\} \ s.t.\ P\left(c\vert\mathbf{x}\right)\leq t\vert+\vert\left\{ c\in\left(\mathcal{Y}\setminus C\right)\right\} \ s.t.\ P\left(c\vert\mathbf{x}\right)\geq t\vert\label{eq:prob_threshold}
\end{equation}
where $\mathcal{Y}$ is the set of all genres. At test time, the XGBoost
classifier and regressor estimate the probabilities of each genre
and the threshold, and output genres with higher probabilities than
the threshold. I use the XGBoost Python Package \cite{xgboost2016package}
to train the XGBoost classifier and regressor. Instead of reinventing
the wheel, I make use of the pretrained 300-dimensional embeddings
trained on part of the Google News corpus for 3 billion words \cite{pretrained2013wordtovec}.

\subsection{Recurent Neural Networks}

Lastly, I train a RNN for this task. The RNN learns not only the embedding
for each word, but also a nonlinear transformation of a sequence of
embedding vectors. Thus, it might learn a more suitable representation
of plot summaries for the genre tagging task. The key assumption in
RNN is the same function can be used to transform current state and
input into a new state. This idea of parameter sharing across time
steps makes it possible to apply the model to examples of different
length.

More specifically, I train a Self-normalizing GRU (see Section~\ref{sec:Related-work})
network with 2 layers, each of which has 128 memory cells. The input
to the network a sequence of 128-dimension vectors, each representing
a word in a plot summary. The network learns vector representations
for a vocabulary of 60,000 words, including the top 55,998 words that
appear at least 5 times in the train dataset, ``UNK'', which represents
other words, and ``EOS'', which stands for end of sequence. Given
the last state $\mathbf{h}_{N}$, a linear transformation is applied
to output a 20-dimension output vector $y$:
\begin{equation}
\mathbf{y}=W\times\mathbf{h}_{N}+\mathbf{b}
\end{equation}

where $W$ is a $128\times20$ matrix and $\mathbf{b}$ is a 20-dimension
bias vector.

Three approaches are considered to make the final predictions using
the output $\mathbf{y}$. In the first approach, each element $y_{c}$
is fed to a logistic sigmoid function to estimate the probability
of the genre $c$:
\begin{equation}
\sigma_{c}\left(y_{c}\right)=\frac{1}{1+\exp\left(-y_{c}\right)}
\end{equation}

The model makes prediction for each genre independently, and predict
genre $c$ if $\sigma_{c}\left(y_{c}\right)$ is larger than 0.50.
This model is called Binary GRU.

In the second approach, $\mathbf{y}$ is interpreted as the rank values
as discussed in Section~\ref{sec:Related-work}; the loss function
is defined in Eq.~\ref{eq:rank_loss_function}, and the threshold
values on the train dataset are estimated following Eq.~\ref{eq:rank_threshold}.
The model following this approach is called Rank GRU.

In last approach, $y$ is fed into a softmax function to estimate
the probability of each genre:
\begin{equation}
P\left(c\vert\mathbf{x}\right)=\frac{\exp\left(y_{c}\right)}{\sum_{k=1}^{20}\exp\left(y_{k}\right)}
\end{equation}

The target probability for each genre $c$ is $\frac{1}{k}$ where
$k$ is the number of correct genres if $c$ is in the set of correct
genres, and is $0$ otherwise. The network is trained to minimize
the cross entropy between the predicted and the target categorical
distribution. The threshold values on the train dataset are estimated
following Eq.~\ref{eq:prob_threshold}. The model following this
approach is called Multinomial GRU.

For both the second and third approaches, an XGBoost regressor is
trained to predict threshold values on the test dataset as described
in~\ref{sub:Word2Vec-+-XGBoost}. Implementation of the GRU network
is in TensorFlow \cite{tensorflow2015whitepaper}.

\section{Experiment\label{sec:Experiment}}

\subsection{Hyperparameter selection}

There is no hyperparameter to tune for the Naive Bayes models. Due
to limited time for experiment and priority, the default setting for
XGBoost is used. For the GRU network, the hyperparameters include
the number of layers, the number of units in each layer, the recurrent
dropout keep rate \cite{semeniuta2016recurrent} and the learning
rate for Adam optimizer \cite{kingma2014adam}. Different values for
each hyperparameter are considered as shown in Table~\ref{tab:hyperparameter_selection}

\begin{table}
\begin{tabular}{cc}
\hline 
Hyperparameters & Values\tabularnewline
\hline 
Number of layers & {[}1, 2, 3{]}\tabularnewline
Number of units & {[}128, 256{]}\tabularnewline
Dropout keep rate & {[}0.5, 0.6, 0.7, 0.8, 0.9{]}\tabularnewline
Learning rate & {[}0.0002, 0.0005, 0.001, 0.003, 0.005, 0.008, 0.01, 0.02, 0.05{]}\tabularnewline
\hline 
\end{tabular}

\caption{Different values considered for hyperparameter selection.\label{tab:hyperparameter_selection}}
\end{table}

To tune the hyperparameters, I divide the dataset into the train,
validation and test sets according to the 70/10/20 proportion. The
hyperparameters that yield the highest softmax loss or rank loss in
the validation set are selected. The optimal settings are 2 for the
number of layers, 128 for the number of units in each layers, 0.8
for the dropout keep rate and 0.01 for the learning rate.

\subsection{Evaluation metrics}

A weak metric is the ``\emph{hit rate}'', proportion of examples
where the model predicts at least one correct genre. However, tagging
all movies with all genres can yield a hit rate of 100\%. To punish
for incorrect predictions, I adopt the Jaccard index, which is defined
as the number of correctly predicted labels divided by the union of
predicted and true labels, $\frac{\vert T\cap P\vert}{\vert T\cup P\vert}$.
To further compare performance of different methods, I calculate the
confusion matrix, accuracy, precision, recall, and F-score for each
genre as well as for all of the test data. These metrics are defined
as:

\begin{align}
Accuracy & =\frac{TP+TN}{TP+TN+FP+FN}\\
Precision & =\frac{TP}{TP+FP}\\
Recall & =\frac{TP}{TP+FN}\\
Fscore & =\frac{2\times Precision\times Recall}{Precision+Recall}
\end{align}

where TP, FP, TN and FN are true positive, false positive, true negative
and false negative respectively.

\subsection{Baseline}

I use two approaches to estimate the baseline. The first approach
is a simple heuristic that predict the top 1 (\emph{drama}), top 2
(\emph{drama} and \emph{comedy}) or top 3 (\emph{drama}, \emph{comedy}
and \emph{action}) most popular genres for all movies, resulting Jaccard
indices of 29.2\%, 28.6\% and 22.8\% and hit rates of 45.3\%, 66.4\%
and 71.3\% respectively. So, tagging all movies to \emph{drama} and
\emph{comedy} is a reasonable baseline. The second approach is to
build 20 binary Naive Bayes model for each genre, using casefolding
and the simple whitespace tokenizer. This method achieves an Jaccard
index of 36.3\%, which is a significant improvement over the heuristic
baseline. The hit rate, however, is only 55.5\%. Additionally, the
F-score in this approach is 0.44.

\subsection{Experiment results}

\begin{figure}
\includegraphics[width=1\textwidth]{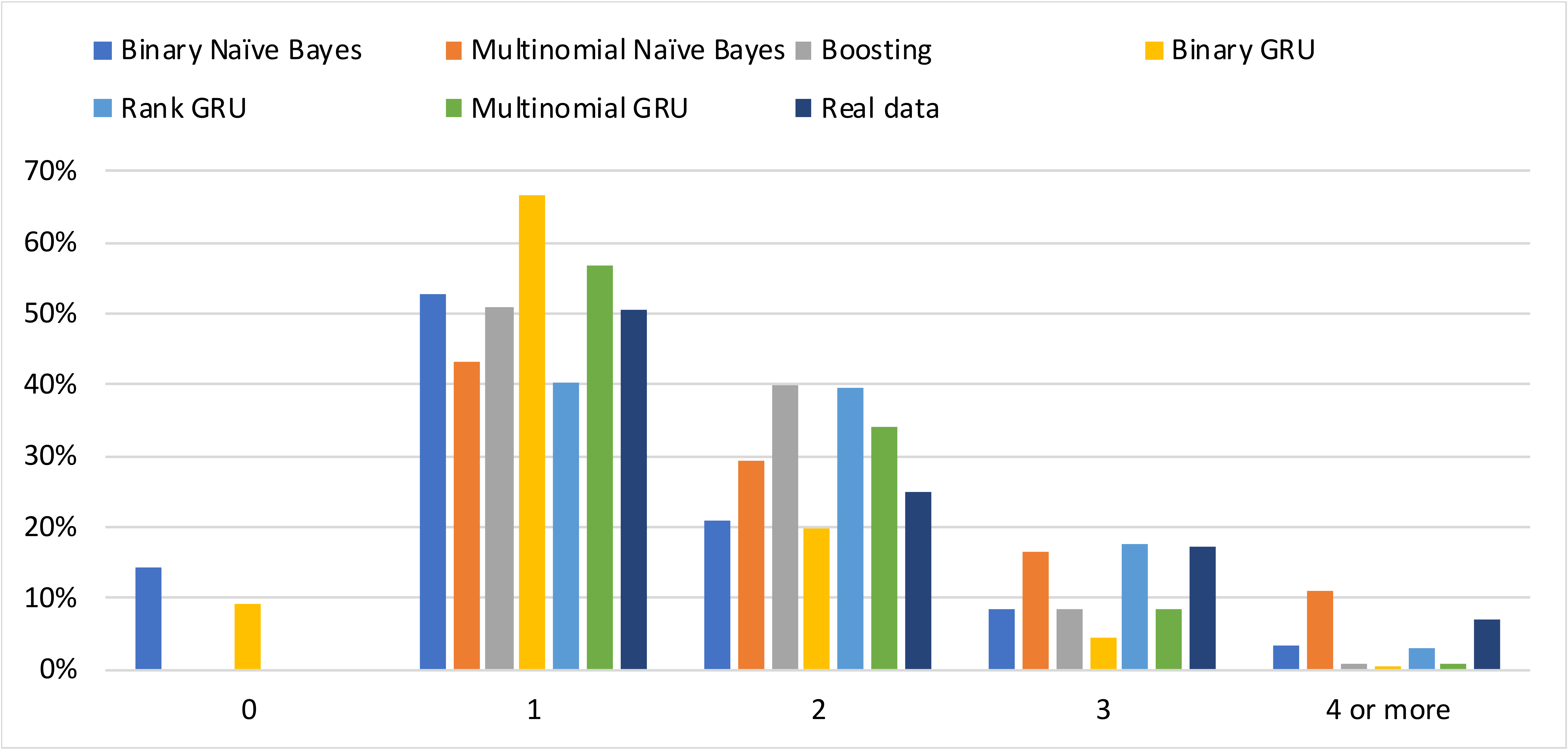}

\caption{Size distribution of the predicted genre sets.\label{fig:Size-distribution}}

\end{figure}

Table~\ref{tab:Summary-of-experiment} summarizes experiment results
of different methods discussed in Section~\ref{sec:Methodology}.
Multinomial GRU achieves the highest Jaccard index of 50.0\% and F-score
of 0.56. Binary-based models such as Binary GRU and Binary Naive Bayes
attain high precision (67.4\% and 60.7\% respectively), while rank-based
models such as Multinomial Naive Bayes, rank GRU and Multinomial GRU
obtain high recall (56.3\%, 56.1\%, and 51.3\% respectively). This
result is plausible as binary-based models make decision for each
genre independently, and return positive prediction only when it is
very confident. Meanwhile, rank-based models make decisions based
on relative rank values (or conditional probabilities) among different
genres, thus output positive prediction more often. This observation
is confirmed by Figure~\ref{fig:Size-distribution}, which shows
that Multinomial Naive Bayes (in orange) and Rank GRU (in light blue)
predict on average more genres per movie. As a result, these models
get the highest hit rates, while Binary GRU and Binary Naive Bayes
get the lowest. Overall, Multinomial GRU and Rank GRU achieve better
balance among different metrics.

\begin{figure}
\includegraphics[width=1\textwidth]{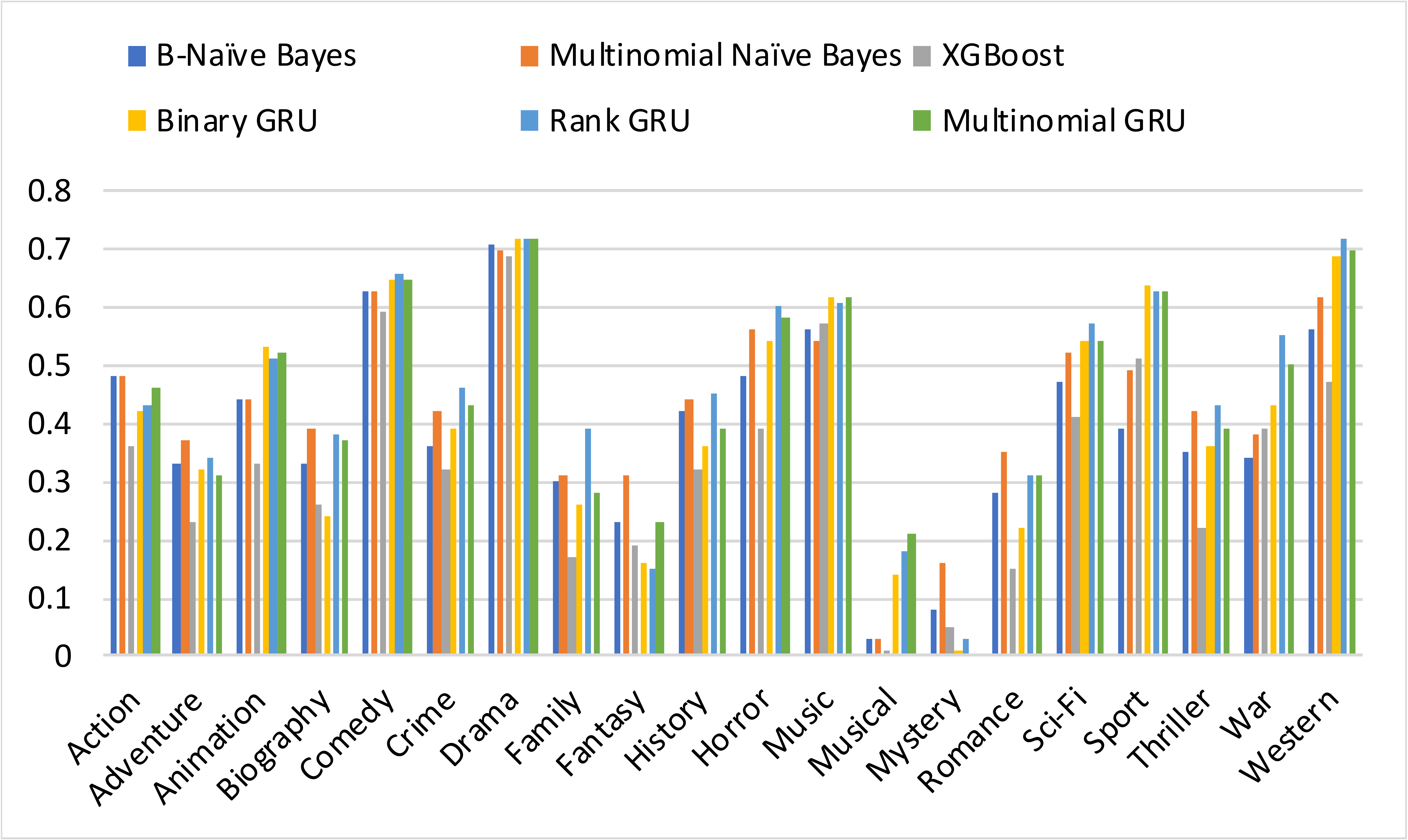}\caption{F-score by genre.\label{fig:F-score-by-genre}}
\end{figure}

Figure~\ref{fig:F-score-by-genre} shows F-score by genre of each
model. It can be seen that GRU models (the last 3 columns in gold,
light blue, and green) consistently attain higher F-score than Naive
Bayes models (the first 2 columns in blue and orange) and XGBoost
(the middle column in gray). XGBoost consistently under-performs across
genres. Tuning hyperparameters might improve this model a little bit,
but this result strongly suggests that taking average of embeddings
vector of all words in a plot summary might be too lossy and does
not yield a good representation. The Naive Bayes models are competitive
with GRU models on the most popular genres such as \emph{Drama} (46\%
of train examples), \emph{Comedy} (27.9\%), \emph{Thriller} (11.2\%),
\emph{Romance} (11.0\%) and \emph{Action} (9.7\%), but under-performs
on less popular genres. So, the GRU models generalize to less popular
genres better. One reason for this behavior is that Naive Bayes takes
the bag-of-word and conditional independence assumption, and the data
is not highly skewed. As a result, Naive Bayes models are biased towards
the most popular genres.

Figure~\ref{fig:Size-distribution} shows the size distribution of
the predicted genre sets by different methods. One problem of the
binary-based models is that many movies are tagged with no genre.
Binary Naive Bayes (in blue) and Binary GRU (in gold) respectively
predict empty set for 15\% and 9\% of test examples. The size distributions
of the genre sets predicted by the rank-based models are closer to
the true distribution.

\begin{table}
\begin{tabular}{c>{\centering}p{1.5cm}>{\centering}p{1.5cm}>{\centering}p{1.5cm}>{\centering}p{1.5cm}>{\centering}p{1.5cm}}
\hline 
Methods & Hit rate & Jaccard & F-score & Precision & Recall\tabularnewline
\hline 
Binary Naive Bayes & 74.2\% & 42.6\% & 0.52 & 60.7\% & 45.0\%\tabularnewline
\rowcolor{even_color}Multinomial Naive Bayes & \textbf{82.9\%} & 46.6\% & 0.53 & 50.7\% & \textbf{56.3\%}\tabularnewline
XGBoost & 74.6\% & 42.9\% & 0.49 & 55.3\% & 44.4\%\tabularnewline
\rowcolor{even_color}Binary GRU & 70.6\% & 46.5\% & 0.53 & \textbf{67.4\%} & 44.0\%\tabularnewline
Rank GRU & 82.7\% & 48.5\% & 0.56 & 56.6\% & 56.1\%\tabularnewline
\rowcolor{even_color}Multinomial GRU & 80.5\% & \textbf{50.0\%} & \textbf{0.56} & 61.0\% & 51.3\%\tabularnewline
\hline 
\end{tabular}

\caption{Summary of experiment results for different methods.\label{tab:Summary-of-experiment}}
\end{table}

The following subsections discuss some models in further details.

\subsubsection{Naive Bayes Models}

As discussed in Section~\ref{sec:Methodology}, I consider two approaches
to apply the Naive Bayes to the movie genre tagging task. The first
approach is to build 20 binary Naive Bayes models for each genre.
This model, when preprocessed with the whitespace tokenizer and casefolding,
is used as the baseline. To better process the data, I apply casefolding,
remove all stop words, and use the NLTK word tokenizer. As a result,
this approach achieves a Jaccard index of 42.6\%, which is a significant
improvement over the whitespace-tokenizer version's 36.3\% score.
Similarly, F-score improves from 0.44 to 0.52.

Using the same word normalization procedure, Multinomial Naive Bayes
attains a Jaccard index of 46.6\% and F score of 0.53.

Figure~\ref{fig:F-score-by-genre} further compares F-score by genre
on the test data. Multinomial Naive Bayes (second columns in orange)
attains much higher F-score than Binary Naive Bayes (first columns
in blue) for less popular genres. The common between the two models
is that they tend to have high recall for very popular genres such
as \emph{drama} or \emph{comedy}, and high precision but low recall
for less popular genres such as \emph{sport}, \emph{war }and \emph{western}
as illustrated in Table~\ref{tab:mNB_precision_recall_fscore}. This
result is plausible given the skewed dataset, which makes the Naive
Bayes models to predict the popular genres more often, and to predict
the less popular genres only when it has strong belief.

\begin{table}
\begin{tabular}{c>{\centering}p{1.5cm}>{\centering}p{1.5cm}>{\centering}p{1.5cm}>{\centering}p{1.5cm}>{\centering}p{1.5cm}}
\hline 
\rowcolor{even_color} & Comedy & Drama & Sport & War & Western\tabularnewline
\hline 
Precision & 54.1\% & 60.2\% & 80.0\% & 65.3\% & 91.1\%\tabularnewline
Recall & 75.3\% & 85.1\% & 35.5\% & 27.1\% & 47.5\%\tabularnewline
F-score & 0.63 & 0.70 & 0.49 & 0.38 & 0.62\tabularnewline
\hline 
\end{tabular}

\caption{The Multinomial Naive Bayes model's precision, recall and F-score
for some genres.\label{tab:mNB_precision_recall_fscore}}
\end{table}

\subsubsection{Multinomial GRU}

This subsection discusses further the Multinomial GRU model. Tab.~\ref{tab:random_tests}
shows Multinomial GRU's predictions on some random test examples.
These few examples illustrate that predicting movie genres is a challenging
task even for human. Plot summaries do not convey all the information
about a movie, and are sometimes ambiguous. The first example is about
a guy being misunderstood by his girlfriend, who becomes upset and
runs away. The correct genre is \emph{romance} but the model predicts
\emph{drama}, which is quite plausible. In the second example, the
model correctly predicts \emph{western}, but incorrectly predicts
\emph{action}, perhaps due to ``\emph{chases and catches}'' or ``\emph{robber}''.
In the third example, the predicted genres are \emph{drama} and \emph{thriller}
while the true genres are \emph{drama} and \emph{comedy}. However,
the plot with details such as ``\emph{stalking}'', ``\emph{murder}''
and ``\emph{more people to kill}'' suggests \emph{thriller} more
than \emph{comedy}. It should be noted that the Jaccard index is only
33.3\% in this case. In the fourth example, there is no sign of \emph{musical
}unless one knows that Frank Sinatra is singer, while \emph{biography}
seems a reasonable choice.

In the fifth example, the model correctly predicts \emph{comedy} and
\emph{family}, but fails to include \emph{fantasy}, \emph{mystery},
\emph{romance}, and \emph{sci-fi}. Of these missed genres, it is possible
to detect \emph{mystery} or even \emph{fantasy} from the plot summary,
but there is no clue to \emph{romance} or \emph{sci-fi}. As the only
sign of comedy in the plot summary is ``\emph{many comical situations}'',
I changed it into ``\emph{many }\emph{\uline{unexpected}}\emph{
situations}'' and the model predicts \emph{family} and \emph{horror}
instead. The probabilities of \emph{family},\emph{ comedy} and \emph{horror}
are changed from 33\%, 27\% and 9\% to 29\%, 13\% and 21\%, respectively.

The sixth and seventh example shows the model's bias towards the most
popular genres such as \emph{drama} or \emph{comedy}. In the sixth
example, the model assigns a probability of 27\% to \emph{drama},
24\% to \emph{horror}, 15\% to \emph{thriller} and 14\% to \emph{mystery
}and selects only \emph{drama} and \emph{horror}, although there is
no information in the plot summary suggesting \emph{drama}. Similarly
in the seventh example, the model assigns a probability of 46\% to
\emph{romance}, 25\% to \emph{comedy} and 11\% to \emph{drama }and
predict \emph{romance} and \emph{comedy}. The correct genre is \emph{romance}
only.

The model is also sensitive to some words in a plot. In the last example,
the model assigns a probability of 22\% to \emph{sci-fi}, 15\% to
\emph{animation}, 11\% to \emph{action}, 11\% to \emph{fantasy}, and
9\% to \emph{horror}, and it predicts \emph{sci-fi} and \emph{animation}.
The model is obviously not biased towards \emph{animation} because
only 6.2\% of train examples belong to the \emph{animation} genre.
It turned out that the culprit is ``\emph{miniaturized}''. When
removing this word from the plot, the model assigns 31\% to \emph{sci-fi},
13\% to \emph{action}, 12\% to \emph{horror}, 7\% to \emph{fantasy}
and 7\% to \emph{animation}, and it correctly predicts only \emph{sci-fi}.
The first example also illustrates this problem. When ``\emph{drives
away with her}'' is changed into ``\emph{falls in love with her}'',
the model changes its prediction from \emph{drama} to \emph{drama}
and \emph{romance}.

Finally, I qualitatively evaluate the model's learned word embeddings
by looking at the nearest words to some random words. The nearest
words to a certain word are those in the vocabulary with embedding
vectors that have the smallest cosine distance with the word's embedding
vector. Table~\ref{tab:nearest_words} shows top 10 nearest words
of 6 nouns, 6 verbs, and 6 adjectives. Nearest words to nouns are
very relevant. The results, however, are mixed for verbs and adjectives.
Words such as \emph{kill}, \emph{love}, \emph{escape} and \emph{cruel}
have relevant nearest neighbors, but \emph{manipulate}, \emph{preserve},
\emph{arrogant},\emph{ }and \emph{beautiful} have totally irrelevant
neighbors. One possible reason for this behavior is that verbs and
adjectives are often used in a wider range of contexts than nouns.
\emph{Hang}'s neighbors such as ``\emph{smothering}'' or ``\emph{re-connect}''
seem related to different meanings of the word. \emph{Miniaturized}
is an interesting example as it appears only 7 times in the train
data, making the context specific. As a result, among its nearest
words are \emph{karaati}, \emph{gnomes}, \emph{alchemist}, \emph{meadow}
and \emph{fairy-tale}, which may explain its influence on the model's
prediction of \emph{animation} in the previously discussed example.

\begin{table}
\begin{tabular}{>{\centering}p{0.7\textwidth}>{\centering}p{0.11\textwidth}>{\centering}p{0.1\textwidth}}
\hline 
Plot Summary & Predictions & Targets\tabularnewline
\hline 
The story is a typical misunderstanding in a relationship. Marco kisses
a woman on the cheek and drives away with her, overseen by his jealous
girlfriend Marie who writes him a letter and runs away to clear her
mind. After hours of looking for her, Marco finally find his raging
girlfriend and tells her who that mysterious woman really is.  & Drama & Romance\tabularnewline
\rowcolor{even_color}Red Ryder (Bill Elliott as Wild Bill Elliott)
chases and catches a bank robber, but the robber's boss, Denver Jack
(Roy Barcroft) has him released by a crooked lawyer, Larry Randall
(Robert Grady). Later, Randall decides he wants to reform and tells
Denver he is quitting. Denver has him framed on a murder charge and
Randall is to be hanged...unless Red Ryder and Little Beaver (Bobby
Blake) can uncover evidence to prove his innocence. & Western, Action & Western\tabularnewline
Pickups, the new film directed by Jamie Thraves, is about a man called
Aidan (conveniently enough, played by Aidan Gillen) who is suffering
from insomnia, back trouble and the breakdown of his marriage. Aidan
finds solace in a number of strangers he picks up, although he's now
concerned someone is stalking him. Work is getting on top of him too,
he murdered a couple of people last week and he still has more people
to kill. & Drama, Thriller & Comedy, Drama\tabularnewline
\rowcolor{even_color}In the 1950s and 1960s 'Frank Sinatra' (qv)
was the head of the infamous \textquotedbl{}Rat Pack\textquotedbl{}.
He, 'Sammy Davis Jr. (I)' (qv), 'Dean Martin (I)' (qv), 'Peter Lawford'
(qv) and 'Joey Bishop (I)' (qv) worked and played together. This film
dramatizes their volatile relationships with each other and the Kennedys,
'Marilyn Monroe' (qv), mobster 'Sam Giancana' (qv), 'Judith Campbell
Exner' (qv) and the FBI. Sinatra helps 'John F. Kennedy' (qv) get
elected in 1960 with a little help from Giancana. Lawford, married
to a Kennedy, is an unhappy go-between. Davis is fighting racism and
insecurity. Campbell is sleeping with both Giancana and JFK who is
also sleeping with Monroe. & Drama, History, Biography  & Drama, Musical\tabularnewline
The father of the Olsson family is responsible for many people's not
being able to receive any TV broadcasts. Therefore, the family escape
to an old castle in the country, to celebrate their Christmas there,
away from all angry people and dangers. However, the three children
soon discover that something's not right in the old castle. It seems
haunted and it bears a centuries old secret. Many comical situations
occur when the children try to investigate the haunting, but the parents
don't understand what's going on. & Comedy, Family & Comedy, Family, Fantasy, Mystery, Romance, Sci-Fi\tabularnewline
\rowcolor{even_color}Kaden Mackenzy, a teenage girl with the ability
to communicate with the dead, stumbles upon a secret that has been
haunting the small town of Peabody, Massachusetts for ages. Kaden's
gift suddenly becomes her curse as these interactions soon unfold
a mystery of a historic and vicious nature. & Drama, Horror & Horror, Thriller\tabularnewline
A barista discovers and experiences a romance in potency between aromas
and sensations in a coffee shop. A coffee barista discovers and wraps
himself in the magic of the sensations with all the atmosphere created
by the combination of the aromas of the coffee; People who frequent
the place unexpectedly experience feelings of romance, and she exposed
in this environment, will soon experience feelings of ecstasy and
unimaginable pleasure. & Comedy, Romance & Romance\tabularnewline
\rowcolor{even_color}The Investigator, a benevolent alien from a
distant galaxy, selects an Earth boy and girl, John and Julie, to
assist him in his mission to make their world a better place. The
pair are miniaturized to assist The Investigator more easily, and
assigned to prevent the theft by Stavros Karanti of a 14th century
masterpiece from a church on Malta. John and Julie are presented with
a car and a boat, scaled to accommodate their miniaturized size, and
set out to thwart Karanti's plans... & Animation, Sci-Fi & Sci-Fi\tabularnewline
\hline 
\end{tabular}

\caption{Multinomial GRU's predictions for some random test examples.\label{tab:random_tests}}

\end{table}

\begin{table}
\begin{tabular}{>{\centering}p{0.12\textwidth}>{\centering}p{0.83\textwidth}}
\hline 
Words & Nearest words (from left to right)\tabularnewline
\hline 
Adventure & ropes, unsupported, roped, paddle, rainforest, superheroines, goblin,
superheroine, shahenshah, huliau\tabularnewline
\rowcolor{even_color}Cigarette & unethical, porn, twenty-something, uneducated, seated, wonderfully,
reevaluate, rekindles, pretense, charm\tabularnewline
Batman & researched, gotham, organization, enemies, samurai, super, alliance,
bhutan, responsibly, politicized\tabularnewline
\rowcolor{even_color}Cowboy & cowboy, marshal, broncho, saloon, rustling, cattlemen, stagecoach,
ranch, outlaws, ranchers, settler\tabularnewline
Earth & earth, spacecraft, orbs, portal, universe, preternatural, species,
activating, quantum, dimension, cosmic\tabularnewline
\rowcolor{even_color}Mother & child, issue, five-year, family, care, familial, caring, mourning,
wait, ferret\tabularnewline
Escape & protector, humbled, tastes, danger, michaelis, exterminated, protect,
barbarians, pitiless, legit\tabularnewline
\rowcolor{even_color}Kill & killing, attacked, lethal, fury, homicide, deadly, ruthless, captor,
shortcut, kills\tabularnewline
Love & attracted, kiss, crush, playboy, him.but, lovers, eun, i.e, woo, eun-ha\tabularnewline
\rowcolor{even_color}Manipulate & prosecutors, slattery, temps, malti, mousy, psychiatry, joking, 1989,
loveless, bombard\tabularnewline
Hang & conditions, re-connect, osman, smothering, heterogeneous, helps, krishna,
posted, stanislav, feels\tabularnewline
\rowcolor{even_color}Preserve & breasts, jayne, operated, spinning, dismantling, guerrero, shell,
gardens, geography, mavis\tabularnewline
Arrogant & sheer, rhodes, zap, stealthily, adversaries, unbeknownst, stereotyped,
elso, fliers, suzy\tabularnewline
\rowcolor{even_color}Beautiful & automaton, rong, wang, radha, floats, repent, chooses, troublemakers,
dreams, stampeded\tabularnewline
Cruel & observation, bothering, consulted, marat, ae, mennonite, repentance,
seward, bribery, viewfinder\tabularnewline
\rowcolor{even_color}Happy & odette, instant, keynes, aficionados, miserably, drowning, shine,
gossip, west, flirts\tabularnewline
Sad & despise, analogy, stems, sang, gauri, obsessing, sympathizes, cabinets,
language, favouring\tabularnewline
\rowcolor{even_color}Miniaturized  & meadow, gladiatorial, dispenser, karaati, gnomes, pricing, can-do,
alchemist, fairy-tale, dogpatch\tabularnewline
\hline 
\end{tabular}

\caption{Examples of top 10 nearest words defined by cosine distance between
the Multinomial GRU model's learned word embeddings.\label{tab:nearest_words}}
\end{table}

\section{Conclusion\label{sec:Conclusion}}

This project explores several Machine Learning methods to predict
movie genres based on plot summaries. This task is very challenging
due to the ambiguity involved in the multi-label classification problem.
For example, predicting \emph{adventure} and \emph{thriller} against
true labels of \emph{adventure} and \emph{action} yields a Jaccard
Index of only 33.3\%.

Word2vec$+$XGBoost performs poorly as the average of embedding vectors
of words in a document proves a weak representation. A future direction
is to apply doc2vec \cite{le2014distributed} to learn richer representations
of documents. Experiments with both Naive Bayes and GRU networks show
that combining a probabilistic classifier with a probability threshold
regressor works better than the k-binary transformation and the rank
methods for the multi-label classification problem. Using a GRU network
as the probabilistic classifier in this approach, the model achieves
impressive performance with a Jaccard Index of 50.0\%, F-score of
0.56 and hit rate of 80.5\%.

There are several potential directions to improve the GRU network.
As 46\% of words in the vocabulary occur fewer than 20 times in the
train data, most word embeddings get only a few weight updates. Using
pretrained embeddings might be better for these words. In addition,
about 75\% of tokens are turned into ``UNK''. As a result, the GRU
network learns the same embedding for these words. Finding a way to
adapt the UNK's embedding based on context might make the network
more powerful. Finally, the data is highly skewed, making the model
biased towards popular genres such as \emph{drama} or \emph{comedy}.
Dealing with this issue would further improve the performance.

\bibliographystyle{plain}

\end{document}